\documentclass[10pt,twocolumn]{article}  

\usepackage{authblk}   
\usepackage{microtype}
\usepackage{graphicx}
\usepackage{subcaption}
\usepackage{booktabs} 
\usepackage{cite}      
\usepackage{tabularx}
\usepackage{abstract}   
\usepackage[square,numbers]{natbib}

\usepackage{times}                       
\usepackage{setspace}                    
\usepackage{lipsum}
\usepackage{etoolbox}

\makeatletter
\AtBeginEnvironment{thebibliography}{%
  \let\href\@secondoftwo 
  \def\url#1{}
  \def\doi#1{}
}
\makeatother

\newtheorem{definition}{Definition}

\usepackage[letterpaper,left=1in,right=0.75in,top=1in,bottom=1in,margin=1in]{geometry}

\newcommand{\onlineref}[1]{\textit{#1}}

\title{Localized Forest Fire Risk Prediction: A Department-Aware Approach for Operational Decision Support}

\author[1]{Nicolas Caron}
\author[1]{Hassan Noura}
\author[1]{Christophe Guyeux}
\author[2]{Benjamin Aynes}

\affil[1]{Université Marie et Louis Pasteur, CNRS, institut FEMTO-ST, F-90000 Belfort, France, France}
\affil[2]{SAD Marketing, Lille, France}

\date{}  

\begin{document}

\maketitle
\thispagestyle{empty}









\begin{abstract}
Forest fire prediction involves estimating the likelihood of fire ignition or related risk levels in a specific area over a defined time period. With climate change intensifying fire behavior and frequency, accurate prediction has become one of the most pressing challenges in Artificial Intelligence (AI). Traditionally, fire ignition is approached as a binary classification task in the literature. However, this formulation oversimplifies the problem, especially from the perspective of end-users such as firefighters. In general, as is the case in France, firefighting units are organized by department, each with its own terrain, climate conditions, and historical experience with fire events. Consequently, fire risk should be modeled in a way that is sensitive to local conditions and does not assume uniform risk across all regions. This knowledge is generally ignore in the current researches on wildfire prediction with AI. This paper proposes a new approach and a new database that tailors fire risk assessment to departmental contexts, offering more actionable and region-specific predictions for operational use. With this, we present the first national-scale AI benchmark for metropolitan France using state-of-the-art AI models on a relatively unexplored dataset. Finally, we offer a summary of important future works that should be taken into account. Supplementary materials are available on github \onlineref{NicolasCaronPro/Localized-Forest-Fire-Risk-Prediction-ADepartment-Aware-Approach-for-Operational-DecisionSupport}.
\end{abstract}


\section{Introduction}
The immediate cost of wildfire suppression, compensation for damaged property, and the aftermath of lost agricultural productivity and tourism put a strain on local economies. For instance, the 2017 wildfires in the Mediterranean region resulted in estimated economic damages exceeding €200 million, according to \onlineref{Le Monde (2017)}. From a human perspective, wildfires in France often lead to evacuations, sometimes involving thousands of residents and tourists. Health risks related to smoke inhalation and post-fire water contamination are also significant concerns~\cite{finlay2012health,chen2021cardiovascular,cascio2018wildland}. Ecologically, these fires can cause irreversible harm to unique ecosystems. Additionally, wildfires contribute to topsoil erosion, threatening agriculture and leading to landslides in hilly terrains. Lastly, the emissions from these fires contribute not only to local air pollution but also to France's national carbon footprint, further complicating efforts to combat climate change. According to~\onlineref{Reporterre (2024)}, more than 4,000 hectares have burned in 2024.


\subsection{Problem formulation}
In this work, the problem of predicting the future forest fire risk is considered. We define the wildfire prediction in definition~\ref{wildfirePredictionDef}.

\begin{definition}
\textbf{Wildfire prediction} : Predict the future risk of wildfire ignition (or a linked-value) \( V \) in a particular area \( A \), for a particular time range \( T \). Knowing a particular set of characteristics \( F \), we can mathematically define the risk of wildfire as:
\[
V(A, T) = R(F(A, T))
\]
where R is a particular set of functions or an algorithm that calculates the risk value.
\label{wildfirePredictionDef}
\end{definition}

\subsection{State of the art}

In the literature, we found publicly available datasets used to address the wildfire prediction problem.

Kondylatos et al. \cite{WildfireDangerPrediction} released a dataset for wildfire hazard prediction in Greece, covering the period 2009–2021 at a daily spatial resolution of approximately $1,\mathrm{km}^2$. The dataset integrates meteorological variables, land cover, and population data. Each file is about 23 GB in size, which raises significant challenges for memory usage and accessibility.

The SeaFire Cube platform~\cite{karasante2023seasfiremultivariateearthdatacube} offers 21 years of global data (2001--2021), 8-day intervals, $0.25^\circ$ resolution. It includes atmospheric, climate, vegetation, socioeconomic, and fire variables (burned area, CO$_2$). Easy to access with tutorials, it suits seasonal prediction but not fine-scale management. Michail et al.~\cite{michail2025firecastnetearthasagraphseasonalprediction} used it with GraphCast~\cite{lam2023graphcastlearningskillfulmediumrange}, reaching global AUPRC $0.64$, but only $0.20$ in Europe, highlighting poor regional generalization and the need for domain-specific models.

The EO4WildFires benchmark~\cite{sykas_2023_7762564} targets severity forecasting (affected area). It combines multi-sensor series (Sentinel-1, Sentinel-2, NASA Power) for 31,730 events across 45 countries (2018--2022), annotated with EFFIS ($\sim$25 GB in one file). It is rich in features (temperature, precipitation, soil moisture, snow), but large volume hampers usability.

Mesogeos \cite{kondylatos2023mesogeosmultipurposedatasetdatadriven} provides a dataset for the Mediterranean region spanning 2006–2022, with a daily spatial resolution of approximately $1,\mathrm{km}^2$ and 27 variables covering meteorology, vegetation, land cover, and human activity. It includes records of ignitions and burned areas larger than 30 ha. Despite the availability of extraction tools and a leaderboard, the dataset’s large volume poses significant challenges. Moreover, excluding smaller fires also removes important signals, such as detection time and response efficiency. Among the evaluated models (LSTM, GTN, and Transformer), LSTM achieved the best performance.

Across all these datasets, the wildfire risk prediction problem is typically treated as a binary classification task (using the predicted probability to generate risk maps).  
Employing a very fine spatial resolution (1 km $\times$ 1 km) allows for a detailed representation of spatial variables (e.g., land cover, vegetation dryness).  
However, this also introduces a strong bias in the evaluation of metrics, since only a tiny fraction of the test samples correspond to fire events (e.g., only 1,228 fire pixels in~\cite{WildfireDangerPrediction} out of several million pixels).  
Furthermore, the aspect of model calibration has rarely been studied and models often replicate known high-risk zones, while it has been proved that accuracy drops at fine scales~\cite{vilar2010model, https://doi.org/10.1002/eap.2316}.
Another limitation is the large memory consumption required, which restricts their use in practice.
Finally, one element not provided by current datasets is information on the organization of firefighting units within each country, even though risk prediction should ideally be aligned with this scale of organization.

\subsection{Contribution}

Our contribution rests on two main points:
\begin{enumerate}
    \item Unlike existing public datasets, ours integrates the administrative structure of French departments, which aligns with the fire service organization and makes predictions more realistic for operational use. We incorporate calendar features (day, holidays, etc.), detailed forest cover information (tree species), refined land cover, and a comprehensive set of fire indices (FWI, Nesterov, Angström). Spatial data are provided in two formats to facilitate categorical handling. All features are aggregated into an $2,\mathrm{km} \times 2,\mathrm{km} \times 1$ day xarray. To manage memory, only one departmental cube is loaded at a time—this increases aggregation time but enables flexible study areas ranging from a single department to the whole of France. We demonstrate usage through fire count prediction with an ordinal classification scheme, though the dataset also supports broader applications in risk management and analysis. Similar studies to~\cite{WildfireDangerPrediction, kondylatos2023mesogeosmultipurposedatasetdatadriven} can likewise be conducted.
    
    \item Binary wildfire danger ignores France’s regional heterogeneity: two ignitions in low-risk Brittany may indicate a crisis, while the same in Mediterranean areas is routine. Standard models flatten these contrasts, trivializing near-certain summer fires in the south and exaggerating sparse events elsewhere. This uniformity misguides resource allocation. The prediction of the number of fires (or of the total burned area) is very random (4 vs. 5 vs. 6 fires), which may also limit the model’s convergence. We thus propose the first national AI benchmark for metropolitan France, replacing binary labels with a region-aware multi-class scheme.
\end{enumerate}

\section{Constructing the database}
In this section, we present the original fire source we used to construct the database along with the original features sources.
\begin{figure*}[h!]
\begin{center}
    \begin{subfigure}[b]{\textwidth}
        \centerline{\includegraphics[width=0.98\textwidth]{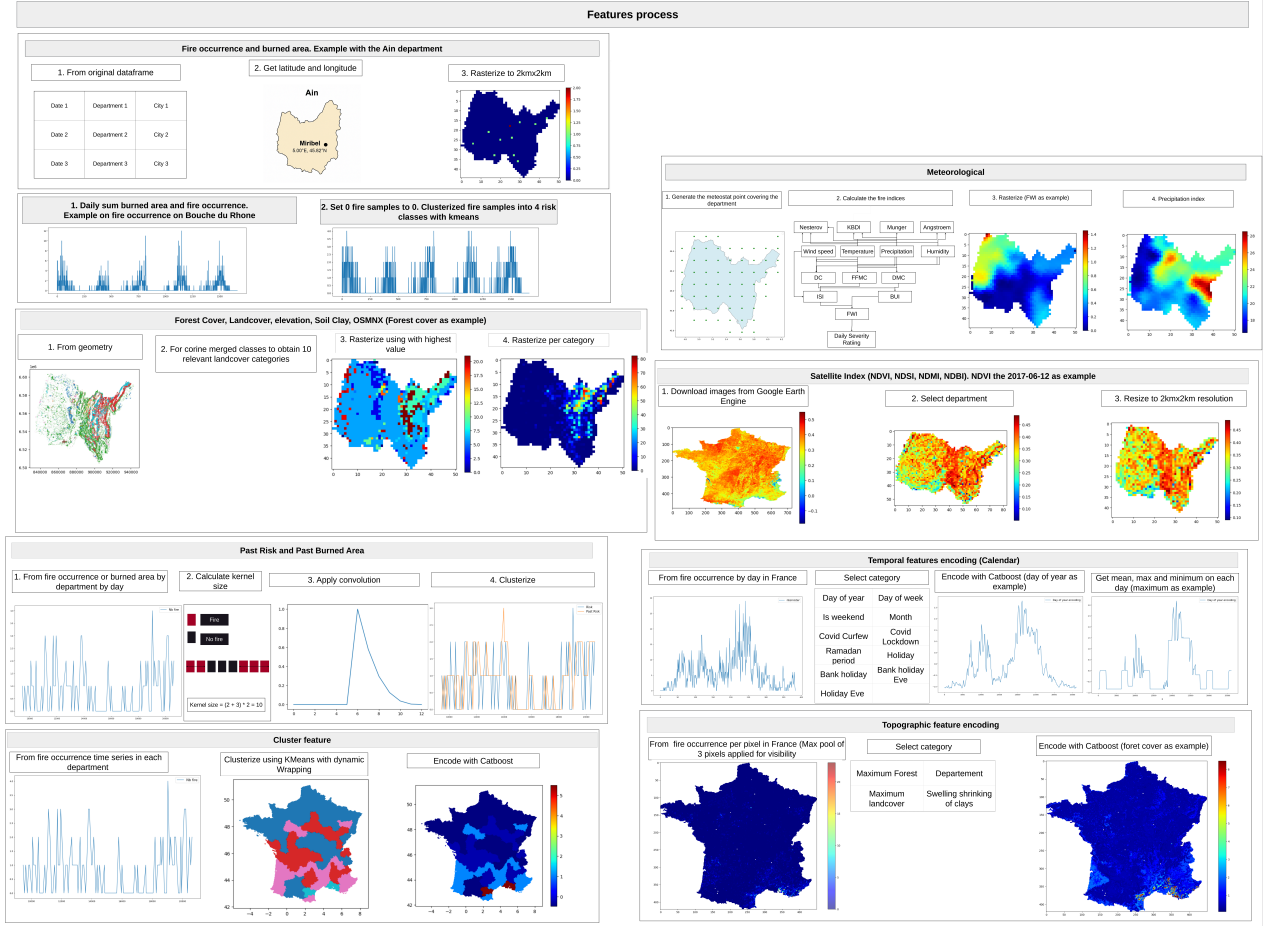}}
    \end{subfigure}
    \hfill 
    \caption{Database construction process applied in this study. Apart from the target-related process, for which we show Bouches-du-Rhône, the department shown is Ain.}
    \label{fig:features_process}
\end{center}
\vspace{+0.5cm}
\end{figure*}

\subsection{Processing Fire occurrence and Burned Area}
\onlineref{BDIFF} is an internet application designed to centralize all data on forest fires across French territory since 2006 and make this information available to the public and state services. Fire locations are referenced by the name of the city where they occurred; coordinates are not provided. Fire are defined, notably, by the BA.
 
Data from June 12, 2017, to December 31, 2024 were collected. The years 2017–2020 and 2022 were used for training, 2021 and 2024 for validation, and 2023 for testing. Due to the severity of fire activity in 2022, it was included in the training set, while a more typical years was selected for validation. We did not include 2024 in the test split but used it for validation, because it was a relatively low-fire year (about 1,600 fires vs. 2,300 in 2023), which would make it a less reliable test year.



City names were used to geolocate fire points and to generate $2,\mathrm{km} \times 2,\mathrm{km}$ rasters for each department. Two daily targets were defined: Fire Occurrence (FO) and Burned Area (BA). Instead of treating these as binary or regression tasks, we reformulated them as multi-class problems by constructing an ordinal five-class signal for both FO and BA using K-Means clustering. Class~0 corresponds to the absence of fire, while positive samples were grouped into four levels (Normal, Medium, High, Extreme). This approach emphasizes typical rather than absolute values, thereby adapting to departmental variability and improving interpretability for firefighting operations. All predictions are made at the daily scale.

\begin{figure}[h!]
\begin{center}
    \begin{subfigure}[b]{0.5\textwidth}
       \includegraphics[width=\columnwidth]{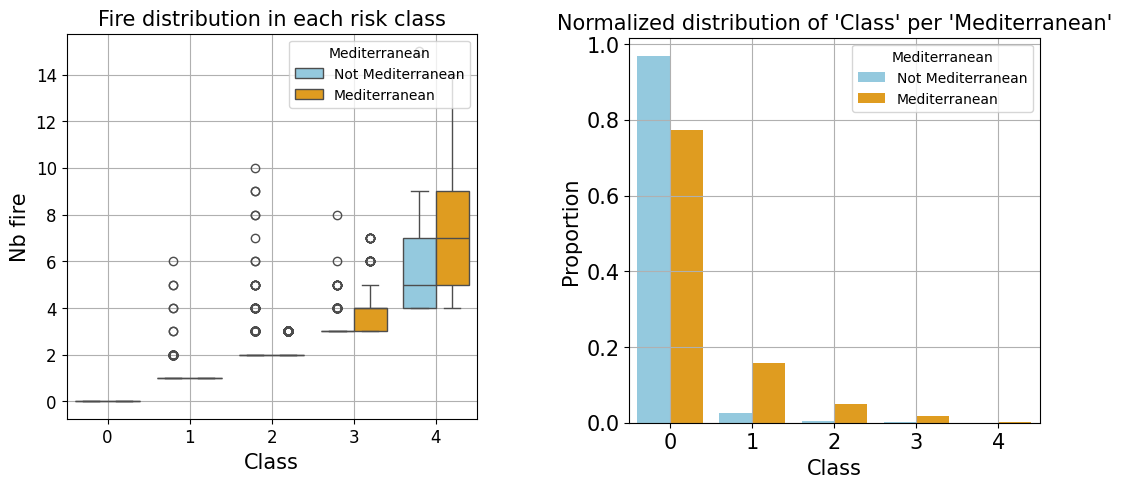}
        \caption*{(a)}
    \end{subfigure}
    \begin{subfigure}[b]{0.5\textwidth}
      \includegraphics[width=\columnwidth]{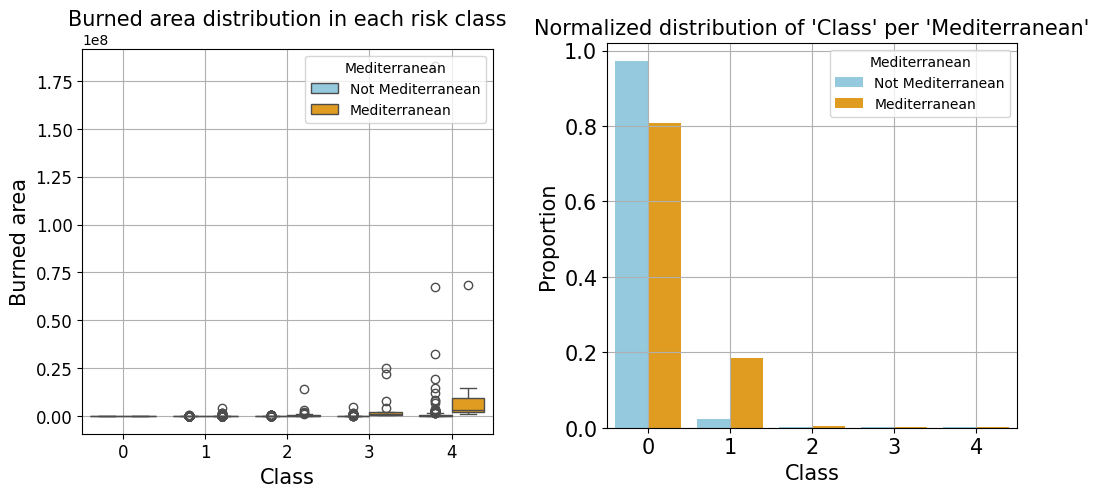}
       \caption*{(b)}
        \vspace{+0.5cm}
    \end{subfigure}
    \caption{Distribution of FO (a) and BA (b) classes risk in the Mediterranean basin and the rest of France. Histograms have been computed relative to the category -Mediterranean (orange) or not (blue)- taking all horizons together.}
    \label{fig:distribution}
\end{center}
\end{figure}

Figure~\ref{fig:distribution} compares class distributions between the Mediterranean basin and the rest of France. Daily FO levels are similar nationwide (except for class 3 and 4, which are concentrated in the Mediterranean). For the BA, class 4 shows high variance, suggesting that extreme events are scattered and department-specific. This supports the idea that our clustering method enhances predictability but may obscure moderate events when extreme cases dominate a department's historical profile. The resulting class distributions show a strong imbalance, with class 0 dominating. Class 1 is the most frequent among positive classes for both targets. Superior classes (2,3 and 4) are underrepresented, specially in BA.

\subsection{Processing Features}

\begin{table*}[!htb]
  \begin{center}
  \caption{Summary of features used in this study. '-' means the same as above.}
  \label{tab:variables}
 \resizebox{\linewidth}{!}{%
\small
  \begin{tabular}{p{4cm}p{2cm}p{2cm}|p{4cm}p{2cm}p{2cm}}
    \toprule
    \textbf{Variables} & \textbf{Frequency} & \textbf{Source} & \textbf{Variables} & \textbf{Frequency} & \textbf{Source} \\
    \midrule
    \multicolumn{3}{l}{\textbf{Meteorological}} & \multicolumn{3}{l}{\textbf{Topographic}} \\
    \midrule
     Temperature & 12h, 16h & Meteostat & Elevation & Static & CourbesDeNiveau (IGN) \\
     Dew Point & - & - & Forest landcover & - & BDForet (IGN) \\
     Precipitation & - & - & Landcover & - & Corine \\
     Wind Direction & - & - & NDVI, NDSI, NDMI, NDBI, NDWI & 15 days & GEE (landsat 1+2) \\
     Wind Speed & - & - & Swelling-shrinking of clays & - & - \\
     Precipitation in Last 24 hours & - & - & & & \\
     Snow height & - & - & & & \\
     Sum of last 7 days rain drop & - & - & & & \\
     Day since last rain & 12h & - & & & \\
     Nesterov & -  & firedanger & & & \\
     Munger & - & - & & & \\
     KBDI & - & - & & & \\
     Angstroem & - & - & & & \\
     BUI, ISI, FFMC, DMC, FWI, & - & - & & & \\
     Daily severity rating & - & - & & & \\
     Precipitation Index last 3, 5, and 9 days & - & Calculated & & & \\
    \midrule
    \multicolumn{3}{l}{\textbf{Socio-Economic}} & \multicolumn{3}{l}{\textbf{Historical}} \\
    \midrule
     Highway & Static & BDRoute (IGN) & Past risk & Daily & Calculated \\
     Population & - & Kontur & Past risk BA & - &  - \\
     Calendar & Daily &  & Cluster & Static &  - \\
     & & & Department & - & - \\
    \midrule
  \end{tabular}
   }
  \end{center}
\end{table*}

The features used in this database fall into four categories (Table~\ref{tab:variables}): \textit{Meteorological}, \textit{Topographic}, \textit{Socio-Economic}, and \textit{Historical}. Most were rasterized into 2 km resolution datacubes for consistency, except for historical data, which were computed at the site level (see Figure~\ref{fig:features_process}).

\textbf{- Meteorological} data were calculated on an $11 \times 11$ grid per department. Fire indices followed standard methods and were reset annually. The precipitation index, based on Chen et al.~\cite{ChenJ}, captures short-term rainfall variation and was computed directly on the 3D raster.

\textbf{- Past risk} and \textbf{BA} features represent prior fire activity, processed using cubic kernel convolution (5 classes) and shifted by one day. The kernel size was based on the average fire sequence duration, with a 3-day inactivity threshold defining sequence boundaries.

\textbf{- Cluster variables} were derived using K-Means clustering with Dynamic Time Warping ~\cite{Muller2007}, grouping departments with similar fire patterns.

\textbf{- Calendar features} capture daily context (e.g., weekdays, holidays, curfews) and were encoded using CatBoost~\cite{NEURIPS2018_14491b75} based on departmental fire totals. Aggregated statistics (mean, sum, min, max) were computed across encoders to represent calendar risk.

\textbf{- Categorical features} were also CatBoost-encoded using fire counts per pixel. Where needed, rasters were also split by subcategories (e.g., primary vs.\ secondary roads) to preserve detail lost at 2~km resolution. NDVI, NDSI, NDBI, NDMI, and NDWI (Normalized Difference Vegetation, Snow, Build-up, Moisture, and Water Index) are satellite indices calculated using Landsat 1 and 2 bands. These indices represent key environmental characteristics such as vegetation density, snow cover, built-up areas, soil moisture, and surface water presence—all of which are relevant for assessing fire risk. The Corine database originally contains 44 land cover classes, but some classes were merged in order to aggregate similar information (such as snow and rock simulate no vegetation area). Therefore, logical reduction was applied to retain only the land cover classes most relevant to forest fires: Urban, transport, forest, natural vegetation, agriculture, grassland, natural non-vegetation area, littoral, water area, and wetland. Code is available in the supplementary materials.

Except for \textbf{Department}, \textbf{Historical}, and \textbf{Calendar} data, all spatial features were aggregated (min, max, mean) by cluster. Finally, all variables—aside from \textit{Past risk} and \textit{Past BA}—were standardized.

Storing all x-array files requires about 150 GB, with each data cube averaging ~1.5 GB. Building all x-arrays can take several days - up to a week. We used a Dell Precision 7780 with 32 GB of RAM and a 13th Gen Intel® Core™ i7-13850HX (28 cores). We do not currently have the rights to publish the xarray files, but the code for downloading and reconstructing them is released on GitHub. Automated downloading of raw data is provided where possible; however, some features (e.g., CORINE and GEE) require authentication. All steps is documented.

\subsection{Features selection}
This process produced a total of 261 features. Those with no variance and those highly correlated with each other (Pearson, Spearman, and Kendall correlation threshold of 0.95) were removed, retaining those with the highest variance. This selection resulted in 162 features. The final feature list is accessible in the supplementary materials. The final dataset is composed of 2,758 unique dates (with 2,192 dates corresponding to a fire in France), across 92 departments (Paris was not include due to lack reliability concerning fire. Corse was not include due to the different vegetation compared to the rest of France), for a total of 253,736 rows.

\section{Models}
The models used are described in detail in table~\ref{tab:models}, description of the deep learning layer parameters, and tree-based configuration is available in the supplementary materials. The data imbalance problem was addressed by testing various proportions of class 0, starting from $5\%$ to $100\%$, with a step of $5\%$. The percentage selected maximize the Intersection over Union on the validation set. Time series models (LSTM, GRU, DilatedCNN) were tested with 10 days sequences. All models were trained with CrossEntropy loss. The learning rate for deep learning models was set to 0.00005. All model parameters were tuned based on validation performance, mainly on horizon 0. The FWI system is based on the classification proposed by Dupire et. al~\cite{Dupire}. Standard deviation has been calculating at each step of the training process with 5 runs.



\begin{table}[h!]
\centering
\renewcommand{\arraystretch}{1.2}
\caption{Models used for the benchmark.}
\resizebox{\columnwidth}{!}{%
\begin{tabular}{|l|p{11cm}|}
\hline
\small
\textbf{Model} & \textbf{Description} \\
\hline
Logistic Regression & A linear model that estimates class probabilities using the logistic function, optimized via cross-entropy. Simple and interpretable but limited for highly non-linear relationships. \\
\hline
XGBoost & An optimized gradient boosting algorithm using second-order derivatives and regularization for efficiency and robustness. Fast and accurate, though often requires careful hyperparameter tuning. \\
\hline
CatBoost & A gradient boosting method on decision trees that handles categorical features natively with ordered encodings. Strong performance on tabular data with minimal tuning. \\
\hline
MLP & A Multi-Layer Perceptron applying three ReLU-activated fully connected layers, outputting class probabilities through a Softmax head. \\
\hline
DilatedCNN (1D) & Stacks dilated 1D convolutions with normalization and dropout before a linear classification head producing class scores and Softmax probabilities from the final time step. \\
\hline
GraphCastGRU & Encodes temporal signals with a GRU, feeds embeddings into \texttt{GraphCast}, and applies linear layers plus Softmax activation for graph-level class predictions. \\
\hline
LSTM & Processes sequences with an LSTM block (optionally layer-normalized), applies dropout and ReLU-activated linear layers, and outputs class probabilities via a Softmax classifier. \\
\hline
GRU & Uses a stacked GRU to summarize temporal dynamics, applies normalization and dropout, then maps the hidden state through linear layers followed by Softmax for classification outputs. \\
\hline
\end{tabular}
\label{tab:models}
}
\end{table}

We evaluate the models' performance on each target using two main metrics: 
\begin{enumerate}
    \item Binary F1 score, which measures the performance of predicting the presence of at least one fire. Additionally, we report the precision and recall scores.
    \item Silva et al.~\cite{8489327} introduce AUOC, a metric that jointly captures classification accuracy and ranking error while accounting for class imbalance and unobserved categories. It is computed by tracing paths along the confusion-matrix diagonal, with a Benefit term rewarding large correct entries and a Penalty term proportional to the distance from the diagonal. A perfect model will give an auoc of 0. AUOC is less penalizing when a fire event is predicted as a nearby but incorrect class (e.g., predicting class 0 instead of 1) and vice versa.
    \item Intersection over Union (IoU), which measures how well the predicted risk aligns with actual risk when an event occurs. IoU is well-suited for multiclass wildfire prediction as it accounts for class uncertainty and preserves class ordinality—predicting class 1 instead of 4 is penalized less than predicting 0. With this metric, a false prediction is penalized proportionally to the class (0 to 4 is worst than 0 to 1 and vice versa).
\end{enumerate}




\section{Results}

Tables~\ref{tab:model_comparison_1_7} present the models performance for predicting FO and BA. Table~\ref{tab:model_comparison_1_7_area} present the models performance for predicting fire risk in each department. This is give by equation~\ref{eq:1} for T being each studied area (here French department), and score a score function.
\newcommand{\scoreop}{\mathrm{score}} 

\begin{equation}
Score\;=\;
\frac{\displaystyle \int_0^T \scoreop(t)\,\mathrm{d}t}
     {\displaystyle \int_0^T \mbox{max\_score}\,\mathrm{d}t}
\label{eq:1}
\end{equation}
This score analyze how well the model is able to predict everywhere. The idea behind this equation is to compare the area under the curve representing each department’s score, normalized by the maximum score (here 1 for each metric). By selecting the departments where at least one fire occurred in 2023, we can evaluate the model’s ability to make predictions within each department. Note that this equation works when the score has a maximum value and this value is the best value possible. For that reason we don't compute the area score for auoc (minimum is better).

\subsection{Model Performance on Global Metrics}
Across all models and tasks, recall values are consistently higher than precision, both in global and area-normalized evaluations. For instance, in FO prediction, CatBoost reaches a recall of 0.51 versus a precision of 0.37, while GRU attains 0.56 recall for only 0.31 precision. This behavior is desirable, specially in low risk region where missing an actual fire is more critical than producing false alarms. We observe that FWI achieves the highest recall with the lowest AUOC, while the other metrics remain very low. This suggests that the metric, in its original implementation, has rather limited relevance for predicting occurrences

\subsection{Model-Specific Observations}
Among all tested models, CatBoost and GraphCastGRU achieve the best F1 scores (both 0.43). GraphCastGRU offers superior recall (0.60 vs. 0.33 precision), a higher IoU (0.25 compared to CatBoost’s 0.24), and better spatial generalization. From the area-normalized evaluation, GraphCastGRU maintains F1 = 0.16, IoU = 0.09 for FO, and IoU = 0.08 for BA.

\subsection{Comparison of FO and BA}
When comparing the average IoU values of Fire Occurrence (FO) and Burned Area (BA), FO appears slightly easier to predict. Global IoU scores for FO range between 0.22–0.24 across models, compared to 0.21–0.23 for BA. This indicates that FO, as a target, is more predictable, whereas BA involves higher uncertainty linked to the dynamics of fire spread. There are several reasons that may explain this phenomenon:  
(1) Burned Area is inherently a more stochastic target compared to occurrence, as it depends not only on ignition but also on spread dynamics influenced by local conditions.  
(2) As observed earlier, the class distribution of Burned Area is more imbalanced than that of Fire Occurrence, leading models to underpredict higher BA classes and thus lowering performance metrics.

\subsection{Binary vs. Multi-class Evaluation}
Binary and multi-class models achieve relatively close F1 scores. For FO, binary CatBoost reaches F1 = 0.13 under area normalization, while multi-class CatBoost scores 0.13 as well. GRU binary and multi-class both yield F1 = 0.16. However, precision-recall trade-offs differ: binary classifiers tend to favor higher recall (e.g., 0.26 for GRU binary) at the cost of lower precision (0.13), while multi-class models remain more balanced (precision around 0.13–0.15, recall around 0.22–0.27).  

This confirms that probability prediction alone is not sufficient at this scale: the ordinal structure of fire risk levels carries valuable information that can be better leveraged. Moreover, both global and area-normalized results show that further improvements require going beyond pure probability outputs.

\begin{figure}[h!]
\begin{center}
       \includegraphics[width=\columnwidth]{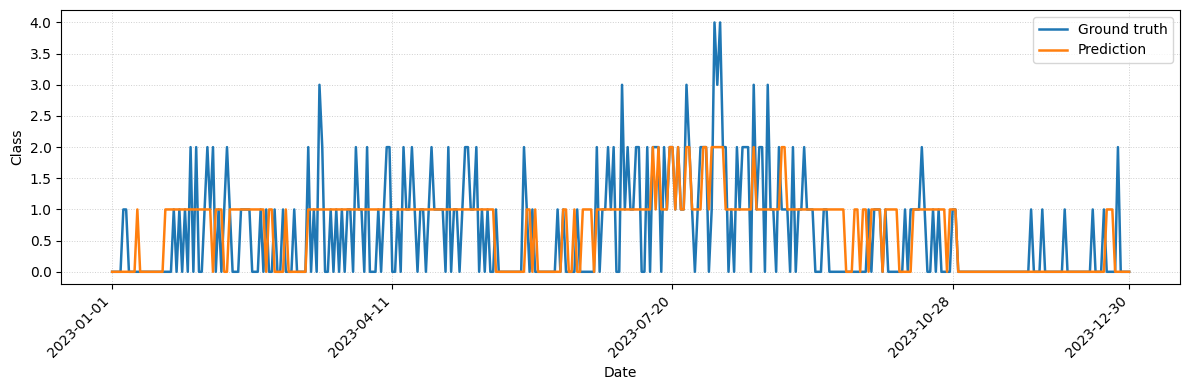}
    \caption{\small GraphCastGRU prediction for Bouches du Rhone in 2023.}
    \label{fig:Prediction}
\end{center}
\end{figure}

Overall, these findings underline the necessity of multi-metric evaluation. Recall-oriented behavior is well aligned with operational needs, but only models such as GraphCastGRU combine high F1 (0.43), strong IoU (0.25), positioning it as the most promising approach for localized fire risk prediction in France. The generalization accross department is a still a main challenge. Figure~\ref{fig:Prediction} shows predictions obtain by GraphCastGRU in Bouches du Rhône. Based on theses predictions we can clearly see the challenging task of predicting highest classes. 

\begin{table}[h!]
\centering
\caption{Model comparison for FO and BA at 0-day ``LR'' stands for Logistic Regression. We do not show FWI performance on BA as it is made for FO. In addition, IoU is not shown for binary models as it is not comparable with multi-classification. Best values are shown in bold.}
\resizebox{\columnwidth}{!}{%
\begin{tabular}{|l|ccccc|cc|}
\hline
Model & \multicolumn{5}{c|}{\textbf{FO}} & \multicolumn{2}{c|}{\textbf{BA}} \\
\cline{2-8}
      & F1 & Prec & Rec & IoU & auoc & IoU & auoc \\
\hline
FWI & 0.17 & 0.10 &  \textbf{0.87} & 0.06 & \textbf{0.619} & - & - \\
LR & 0.39 ± 0.01 & 0.28 ± 0.00 & 0.65 ± 0.01 & 0.23 ± 0.00 & 0.66 ± 0.00 & 0.24 ± 0.00 & 0.71 ± 0.00 \\
catboost & \textbf{0.43 ± 0.00} & \textbf{0.37 ± 0.00} & 0.51 ± 0.00 & 0.24 ± 0.00 & 0.72 ± 0.00 & \textbf{0.25 ± 0.00} & 0.71 ± 0.00 \\
xgboost & 0.42 ± 0.01 & 0.32 ± 0.01 & 0.60 ± 0.01 & 0.24 ± 0.00 & 0.70 ± 0.00 & \textbf{0.25 ± 0.00} & 0.71 ± 0.01 \\
GRU & 0.40 ± 0.01 & 0.31 ± 0.01 & 0.56 ± 0.04 & 0.23 ± 0.00 & 0.70 ± 0.01 & 0.23 ± 0.01 & 0.71 ± 0.01 \\
LSTM & 0.40 ± 0.01 & 0.29 ± 0.01 & 0.61 ± 0.01 & 0.23 ± 0.00 & 0.69 ± 0.01 & 0.23 ± 0.01 & 0.70 ± 0.01 \\
MLP & 0.39 ± 0.01 & 0.31 ± 0.02 & 0.52 ± 0.02 & 0.22 ± 0.01 & 0.71 ± 0.01 & 0.21 ± 0.01 & \textbf{0.69 ± 0.00} \\
DilatedCNN & 0.35 ± 0.01 & 0.24 ± 0.01 & 0.61 ± 0.02 & 0.20 ± 0.01 & 0.70 ± 0.00 & 0.21 ± 0.01 & 0.71 ± 0.01 \\
graphCastGRU & \textbf{0.43 ± 0.01} & 0.33 ± 0.02 & 0.60 ± 0.05 & \textbf{0.25 ± 0.00} & 0.70 ± 0.01 & \textbf{0.25 ± 0.01} & 0.71 ± 0.01 \\
catboost Binary & \textbf{0.43 ± 0.00} & \textbf{0.37 ± 0.01} & 0.50 ± 0.01 & - & - & - & - \\
GRU Binary & 0.40 ± 0.01 & 0.30 ± 0.01 & 0.62 ± 0.01 & - & - & - & - \\
NetMLP Binary & 0.39 ± 0.01 & 0.30 ± 0.01 & 0.55 ± 0.02 & - & - & - & - \\
\hline
\end{tabular}
}
\vspace{0.5cm}
\label{tab:model_comparison_1_7}
\end{table}

\begin{table}[h!]
\centering
\caption{Area model comparison for FO and BA at 0-day prediction.}
\resizebox{\columnwidth}{!}{%
\begin{tabular}{|l|cccc|c|}
\hline
Model & \multicolumn{4}{c|}{\textbf{FO}} & \multicolumn{1}{c|}{\textbf{BA}} \\
\cline{2-6}
      & F1 & Prec & Rec  & IoU & IoU \\
\hline
FWI & 0.14 & 0.08 & \textbf{0.83} & 0.058 & - \\
LR & 0.15 ± 0.01 & 0.14 ± 0.01 & 0.29 ± 0.01 & \textbf{0.09 ± 0.00} & 0.08 ± 0.00 \\
catboost & 0.13 ± 0.00 & 0.13 ± 0.00 & 0.18 ± 0.00 & 0.07 ± 0.00 & 0.08 ± 0.00 \\
xgboost & 0.15 ± 0.00 & 0.13 ± 0.01 & 0.24 ± 0.00 & \textbf{0.09 ± 0.00} & 0.08 ± 0.01 \\
GRU & \textbf{0.16 ± 0.02} & 0.13 ± 0.02 & 0.24 ± 0.04 &\textbf{0.09 ± 0.00} & 0.08 ± 0.00 \\
LSTM & \textbf{0.16 ± 0.01} & 0.13 ± 0.01 & 0.27 ± 0.01 & \textbf{0.09 ± 0.00} & \textbf{0.09 ± 0.01} \\
MLP & 0.15 ± 0.01 & \textbf{0.15 ± 0.01} & 0.22 ± 0.02 & 0.08 ± 0.01 & \textbf{0.09 ± 0.00} \\
DilatedCNN & \textbf{0.16 ± 0.01} & 0.12 ± 0.01 & 0.31 ± 0.03 & \textbf{0.09 ± 0.01} & \textbf{0.09 ± 0.01} \\
graphCastGRU & \textbf{0.16 ± 0.00} & 0.14 ± 0.01 & 0.23 ± 0.04 & \textbf{0.09 ± 0.00} & 0.08 ± 0.01 \\
catboost Binary & 0.13 ± 0.01 & 0.13 ± 0.00 & 0.18 ± 0.00 & - & - \\
GRU Binary & \textbf{0.16 ± 0.00} & 0.13 ± 0.01 & 0.26 ± 0.01 & - & - \\
NetMLP Binary & \textbf{0.16 ± 0.01} & \textbf{0.15 ± 0.02} & 0.24 ± 0.01 & - & - \\
\hline
\end{tabular}
}
\vspace{0.5cm}
\label{tab:model_comparison_1_7_area}
\end{table}



\section{Features importance}\label{fi}

We investigated feature importance by computing SHAP values on the multi-class CatBoost model. Figure~\ref{fig:fet} shows the top 15 features for FO. A corresponding figure for BA prediction is provided in the supplementary materials.

Fire prediction is primarily driven by historical and short-term signals. Encoded temporal variables, spatial clustering, and immediate weather conditions dominate the model’s decision-making. Fire danger indices, relative humidity, and vegetation-specific factors—such as the presence of pine trees—also contribute meaningfully.

Historical features are strong predictors in historically active regions but can hurt generalization in low-risk or emerging areas. While they help the models to easily detect the seasonal trend of fire, their static nature biases the model toward past patterns, potentially missing new fire-prone zones.

\begin{figure}[h!]
\begin{center}
    \begin{subfigure}[b]{0.5\textwidth}
        \centerline{\includegraphics[width=\columnwidth]{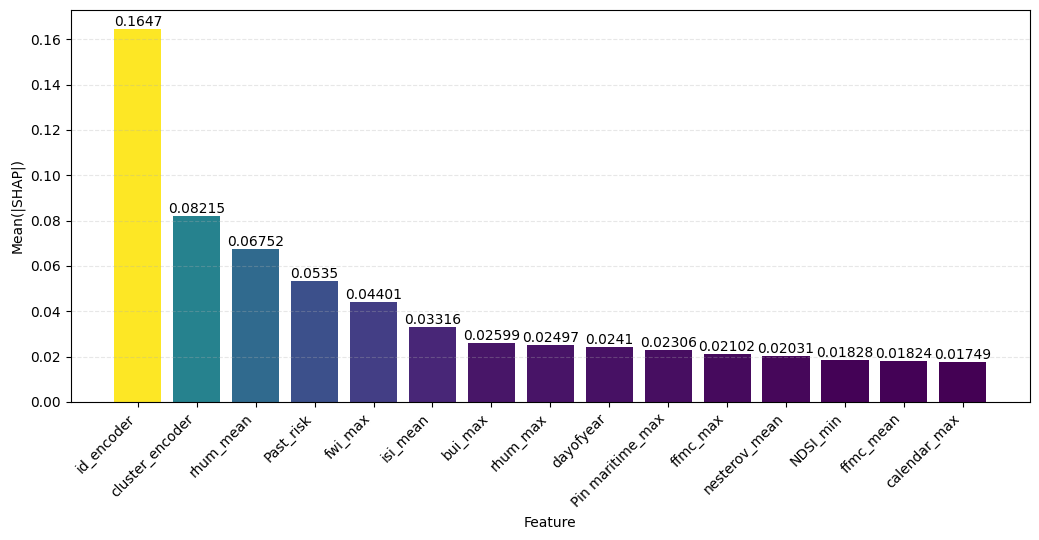}}
    \end{subfigure}
    \hfill
    \caption{Top 15 features computed on multi-classification Catboost model on different time horizons for FO. ID encoder corresponds to Department ID}
    \label{fig:fet}
\end{center}
\end{figure}

\section{Discussion}~\label{futurework}

In this section, we outline several possible directions for future work aimed at improving current model performance.

\paragraph{- Clustering}  
Current risk classes are based solely on historical FOs and BA at the departmental level. Incorporating seasonal data and clustering similar regions may help capture risk dynamics that go beyond purely historical patterns.

\paragraph{- Ordinal Classification Training}  
Risk levels are inherently ordered, yet current models treat them as independent categories. Future work could leverage ordinal-aware loss functions (e.g., Kappa loss) to exploit this structure and better handle class imbalance.

\paragraph{- Federated Learning}  
A single global model may overlook important regional patterns. Federated learning would allow each region or cluster to train a specialized model while still benefiting from shared knowledge to enhance generalization. This approach would also facilitate the inclusion of overseas departments and Corsica, which have been excluded thus far due to their significantly different vegetation and meteorological conditions compared to mainland France.

\paragraph{- Filtering Features}  
Our analysis of feature importance revealed that historical variables are strong predictors in regions with frequent past fire activity. While these features help the model capture seasonal fire trends, their static nature may bias predictions toward past patterns and overlook emerging fire-prone areas. Reducing their influence during training could help the model prioritize more transferable features. We also acknowledge that 2D convolutional networks such as ResNet and ConvLSTM were not included in our benchmark. Although effective for spatial tasks like fine-scale map prediction, their high computational cost made them unsuitable for our department-scale feature set. We plan to explore 2D CNNs (after dimensionality reduction) in future work.

\section{Conclusion}

In this work, we introduced a new nationwide database for wildfire risk prediction, specifically aligned with the departmental organization of French firefighting services and enriched with local variables such as calendar features. While this dataset can serve broader research domains, we proposed the first benchmark for predicting both fire occurrence and burned area using an ordinal classification scheme, a novel approach in this context. Our experiments highlight three key findings: (1) multi-class prediction provides a clear advantage over binary formulations at this scale, (2) GraphCastGRU achieves the best overall performance across metrics which corroborate with findings of~\cite{michail2025firecastnetearthasagraphseasonalprediction}, and (3) the current ordinal scheme is less suitable for burned area prediction. Future directions include exploring ordinal-aware loss functions, applying federated learning to assess generalization, and testing more computationally demanding architectures such as CNNs and Transformers to better capture spatial and temporal dependencies.

\bibliography{main.bib}
\bibliographystyle{unsrtnat}  

\end{document}